\documentclass[letterpaper]{article}
\usepackage{uai2020}
\usepackage[margin=1in]{geometry}

\usepackage{times}

\title{Safe Reinforcement Learning of Control-Affine Systems with \\ Vertex Networks}

\author{Liyuan Zheng, Yuanyuan Shi, Lillian J. Ratliff, and Baosen Zhang \\
Department of Electrical \& Computer Engineering, University of Washington \\
{\tt\small \{liyuanz8,yyshi,ratliffl,zhangbao\}@uw.edu}}

\usepackage{amsmath,amsthm,amssymb}
\usepackage{graphicx}

\newtheorem{thm}{Theorem}[section]
\newtheorem{example}{Example}

\newtheorem{proposition}[thm]{Proposition}
\newcommand{\x}{\mathbf{x}}
\newcommand{\bu}{\mathbf{u}}
\newcommand{\bd}{\mathbf}

\begin{document}

\maketitle

\begin{abstract}
	This paper focuses on finding reinforcement learning policies for control systems with hard state and action constraints. Despite its success in many domains, reinforcement learning is challenging to apply to problems with hard constraints, especially if both the state variables and actions are constrained. Previous works seeking to ensure constraint satisfaction, or safety, have focused on adding a projection step to a learned policy. Yet, this approach requires solving an optimization problem at every policy execution step, which can lead to significant computational costs.  
	
	To tackle this problem, this paper proposes a new approach, termed Vertex Networks (VNs), with guarantees on safety during exploration and on learned control policies by incorporating the safety constraints into the policy network architecture. Leveraging the geometric property that all points within a convex set can be represented as the convex combination of its vertices, the proposed algorithm first learns the convex combination weights and then uses these weights along with the pre-calculated  vertices to output an action. The output action is guaranteed to be safe by construction. Numerical examples illustrate that the proposed VN algorithm outperforms vanilla reinforcement learning in a variety of benchmark control tasks.  
\end{abstract}

\section{Introduction}
Over the last couple of years, reinforcement learning (RL) algorithms have yielded impressive results on a variety of applications. These successes include playing video games with super-human performance~\cite{mnih2015human}, robot locomotion and manipulation~\cite{lillicrap2015continuous,levine2016end}, autonomous vehicles~\cite{sallab2017deep}, and many benchmark continuous control tasks~\cite{duan2016benchmarking}.

In RL, an agent learns to make sequential decisions by interacting with the environment, gradually improving its performance at the task as learning progresses. Policy optimization algorithms \cite{lillicrap2015continuous,schulman2015trust} for RL assume that agents are free to explore any behavior during learning, so long as it leads to performance improvement. However, in many real-world applications, there is often additional safety constraints, or specifications that lead to constraints, on the learning problem. For instance, a robot arm should prevent some behaviors that could cause it to damage itself or the objects around it, and autonomous vehicles must avoid crashing into others while navigating \cite{garcia2015comprehensive}. 


In real-world applications such as the above, constraints are an integral part of the problem description, and maintaining constraint satisfaction during learning is critical (i.e., these are hard constraints). Therefore, in this work, our goal is to maintain constraint satisfaction at each step throughout the whole learning process. This problem is sometimes called the \emph{safe exploration} problem \cite{amodei2016concrete,wachi2018safe}. In particular, we define safety as remaining within some pre-specified polytope constraints on both states and actions. Correspondingly, the action we take at each step should result in a state that in the safety set.

In the safe exploration literature, the projection technique is often leveraged to maintain safety during exploration \cite{cheng2019end,dalal2018safe}. Specifically, at each step, an action is suggested by an unconstrained policy optimization algorithm, and then is projected into the safety region. However, this projection step either involves solving a computationally expensive optimization problem online \cite{cheng2019end}, or has strict assumptions such as allowing for only one of the half-spaces to be violated \cite{dalal2018safe}. More over, if real-time optimization is allowed by the application, then it is often more advantageous to solve a model predictive control problem than to ask for a policy learned by RL. 

To alleviate the limitation, we proposed Vertex Networks (VNs), where we encode the safety constraints into the policy via neural network architecture design. In VNs, we compute the vertices of the safety region at each time step and design the action to be the convex combination of those vertices, allowing policy optimization algorithms to explore only in the safe region.

The contributions of this work can be briefly summarized as follows: (1) To the best of our knowledge, this is the first attempt to encode safety constraints into policies by explicit network design. (2) In simulation, the proposed approach achieves good performance while maintaining constraint satisfaction. 

\section{Related work} 

In safe RL, safety in expectation is a widely used criterion \cite{altman1999constrained,yu2019convergent}. In recent literature, policy optimization algorithms have been proposed as a means to learn a policy for a continuous Markov decision process (MDP). Two  state-of-the-art exemplars in terms of performance are the Lagrangian-based actor-critic algorithm \cite{bhatnagar2012online,chow2017risk} and Constrained Policy Optimization (CPO) \cite{achiam2017constrained}. However, for these methods, constraint satisfaction can only be guaranteed in expectation. In a safety critical environment, this is not sufficient since even if safety is guaranteed in expectation, there is still a non-zero probability that unsafe trajectories will be generated by the controller.


Safe exploration, on the other hand, requires constraint satisfaction at each steps. Recent approaches \cite{sui2015safe,wachi2018safe} model safety as unknown functions, proposed algorithm that trades off between exploring the safety function and reward function. However, their approaches require solving MDPs or constrained optimizations to obtain policy in each exploration iteration, which are less efficient than the policy optimization algorithm leveraged in our approach. 


Literature that uses policy optimization algorithm in safe exploration is closely related to our work. Among those, projection technique often used for maintaining safety. In \cite{dalal2018safe}, the safety set is defined in terms of half-space constraints on the action. A policy optimization algorithm---in particular, deep deterministic policy gradient (DDPG)~\cite{lillicrap2015continuous}---is leveraged to generate an action which is then projected into the safe set. Imposing that  only one half-space constraint can be violated, the projection optimization problem can be solved in closed form. In \cite{cheng2019end}, this projection step is solved as a quadratic program, based on   confidence intervals for the approximation of the system dynamics, modeled as Gaussian processes. 

It is possible to integrate a projection step into the policy network. Indeed, using the methodology provided in \cite{amos2017optnet}, one can leverage a policy optimization algorithm to train the policy network in an end-to-end fashion. However, the feed forward computation of the policy network in this case is computationally expensive as it involves solving an optimization problem before executing every single action. One might prefer to solve a model predictive control problem instead if solving the optimization problem online is involved. Instead of integrating the projection step, we propose VNs which leverage the convex combination of vertices to enforce safety.

\section{Model Setup}


Consider a discrete time affine control system in which the system evolves according to
\begin{equation} \label{eqn:affine}
\x_{t+1} = f(\x_t)+ H(\x_t) \bu_t,
\end{equation}
where $\x \in \mathbb{R}^n$, $\bu \in \mathbb{R}^m$, and $f$ and $H$ are known functions of appropriate dimensions. Our goal is to minimize a cost over time horizon $T$, subject to safety constraints on $\x$ and actuator constraints on $\bu$:
\begin{subequations}
	\label{eqn:opt}
	\begin{align}
	\min_{\bu} \; & \sum_{t=1}^T C(\x_t,\bu_t) \label{eqn:cost}\\
	\mbox{s.t. } &\x_{t+1} = f(\x_t)+ H(\x_t) \bu_t \label{eqn:state_dyn}\\
	& \x_t \in \mathcal{X} \label{eqn:state_constr}\\
	& \bu_t \in \mathcal{U}, \label{eqn:action_constr}
	\end{align}
\end{subequations}
where $\mathcal{X}$ and $\mathcal{U}$ are \emph{convex polytopes}. A convex polytope can be defined as an intersection of linear inequalities (half-space representation) or equivalently as a convex combination of a finite number of points (convex-hull representation)~\cite{grunbaum2013convex}. This type of constraints are widely used in theory and practice---for example, see~\cite{blanchini2008set} and the references within. 

The goal of safe RL is to find an optimal feedback controller $\bu_t = \pi_{\theta}(\x_t)$, that minimize the overall system cost~\eqref{eqn:cost} while satisfies the safety constraints~\eqref{eqn:state_constr} and the actuator constraints~\eqref{eqn:action_constr}. Solving~\eqref{eqn:opt} is a difficult task, even for linear systems with only the actuator constraints, except for a class of systems where analytic solutions can be found~\cite{gokcek2001lqr}. Therefore, RL (and its different variants) have been proposed to search for a feedback controller.  

Numerous learning approaches have been adopted to solve the problem when the constraints  \eqref{eqn:state_constr} and \eqref{eqn:action_constr} are not present. However, there are considerably less successful applications of RL to problems with hard constraints. One such approach is the two-stage method used in~\cite{dalal2018safe}. The first step is to simply train a policy that solves the problem in~\eqref{eqn:opt} without the constraints on state nor the action. To enforce the constraints, a projection step is solved, where the action determined by the unconstrained policy is projected into the constraint sets. 

This two-step process is referred to as safe exploration in~\cite{dalal2018safe}, since it leverages the fact that RL algorithms explore the action space while the projection satisfies the hard constraints. However, this approach has two drawbacks that we address in the current paper. Firstly, the projection step itself requires an optimization problem to be solved. This step could be computationally expensive. More fundamentally, it brings the question of why not directly solve \eqref{eqn:opt} as a model predictive control problem, since online optimization needs to be used either way. Secondly, decoupling the policy and projection steps may lead to solutions that significantly deviate from the original unconstrained policy. To overcome these challenges, we propose a novel vertex policy network that encodes the \emph{geometry} of the constraints into the network architecture, and train it in an end-to-end way. We will discuss the proposed vertex policy network framework in detail in the next section.

\section{Vertex Policy Network}
The key idea of our proposed VN is using a basic fact of the geometry of a convex polytope. Given a bounded convex polytope $\mathcal{P}$, it is always possible to find a finite number of \emph{vertices} such that the convex hull is $\mathcal{P}$. In addition, there is no smaller set of points whose convex hull forms $\mathcal{P}$~\cite{grunbaum2013convex}. Then, the next proposition follows directly.
\begin{proposition}\label{lem:vertices} 
	Let $\mathcal{P}$ be a convex polytope with vertices $P_1,\dots,P_N$. For every point $\bd p \in \mathcal{P}$, there exists $\lambda_1,\dots,\lambda_N$, such that 
	\begin{equation*}
	\bd p= \lambda_1 P_1+\dots+\lambda_N P_N,
	\end{equation*}
	where $\lambda_i \geq 0, \forall i$ and $\lambda_1+\dots+\lambda_N=1$. 
\end{proposition} 
The preceding proposition implies that we can search for the set of weights $\lambda_i$'s instead of directly finding a point inside polytope. 

Proposition~\ref{lem:vertices} can be applied to find a feedback control policy. Since both the constraint sets $\mathcal{X}$ and $\mathcal{U}$ are convex polytopes, the control action at each timestep must also live in a convex polytope. If its vertices are known, the output of a policy can be the weights $\lambda_i$'s. 
The benefit of having the weights as the output is threefold. Firstly, it is much easier to normalize a set of real numbers to be all positive and to sum to unity (the probability simplex) than to project into an arbitrary polytope. For this paper, we use a softmax layer. Secondly, this approach allows us to fully explore the interior of the feasible space, where projections could be biased towards the boundary of the set. Thirdly, we are able to use standard policy gradient training techniques.  

\begin{figure}[h]
	\begin{center}
		\includegraphics[width=0.9\columnwidth]{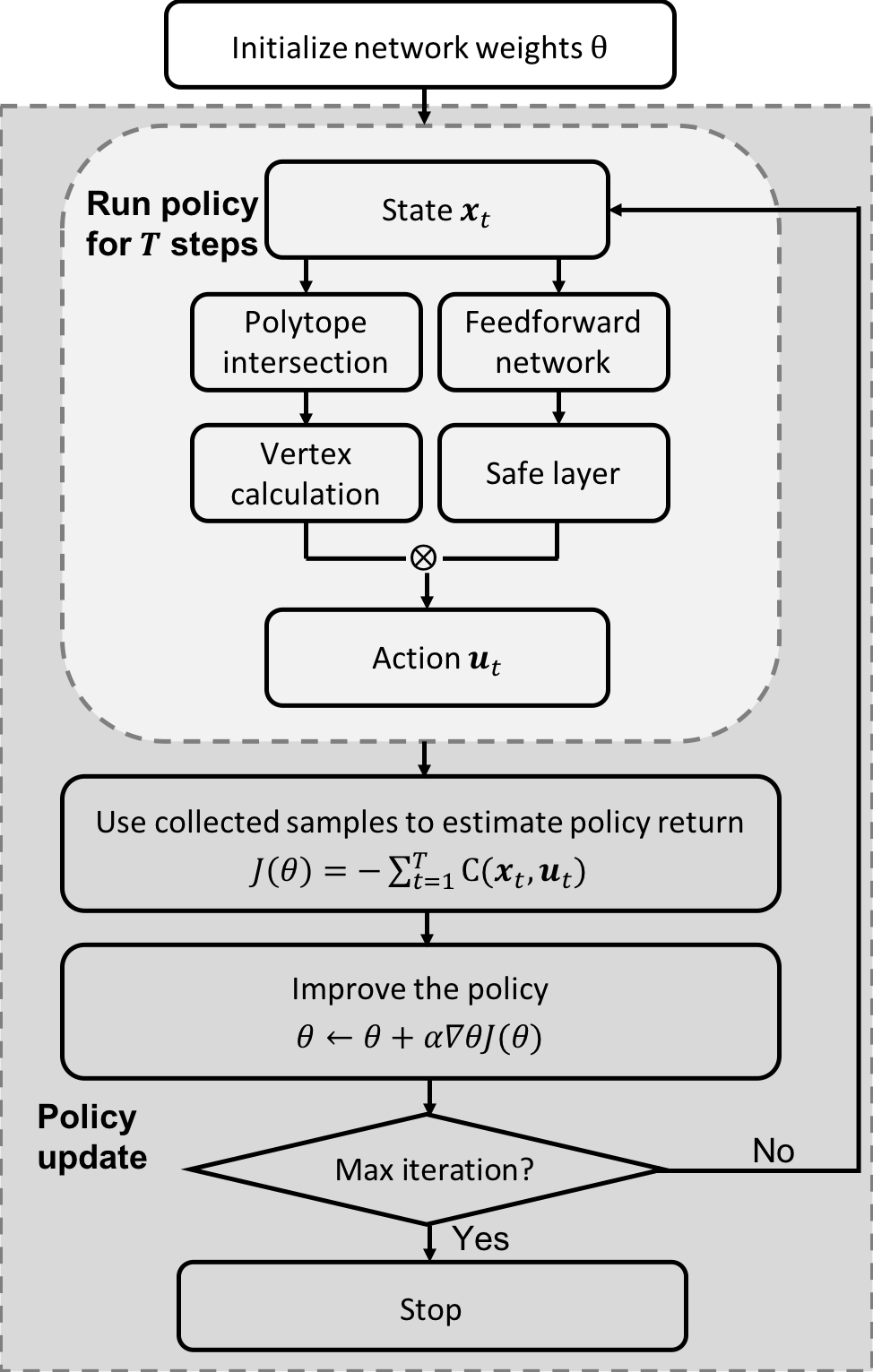}
	\end{center}
	\caption{Flowchart of the proposed VN framework.}
	\label{fig:flowchart}
\end{figure}

In particular, we use DDPG as the policy evaluation and update algorithm, where the policy is a neural network parameterized by $\theta$ and updated by 
\begin{equation}\label{eq:policy_update}
\theta \leftarrow \theta + \alpha \nabla_{\theta} J(\theta).
\end{equation}
where $J(\theta)$ is the expected return using the current policy and is defined by
\begin{equation}
J(\theta) = \mathbb{E}\left[\sum_{t=1}^{T} -C(\x_t, \bu_t)\right].
\end{equation}
We approximate $J(\theta)$ by $-\frac{1}{N} \sum_{i=1}^{N} \sum_{t=1}^{T} C(\x_{i, t}, \bu_{i, t})$, where $N$ are the number of sampled trajectories generated by running the current policy $\pi_{\theta}(\bu_t|\x_t)$ and $T$ is the trajectory length.
The overall algorithm procedure of the proposed VN framework is provided in Fig~\ref{fig:flowchart}. 

Below, we discuss the two major components of VN in detail: 1) the safety region and vertex calculation, and 2) the neural network architecture design for the safe layer.

\subsection{Evolution of the Action Constraint Set}
We require that at each time step the states of the system stay in the set $\mathcal{X}$, and the control actions at constrained to be in the set $\mathcal{U}$. As stated earlier, we assume $\mathcal{X}$ and $\mathcal{U}$ to be convex polytopes. The main algorithmic challenge comes from the need to repeatedly intersect translated versions of these polytopes. 
To be concrete, suppose we are given $\x_t$. Then for the next step, we require that $\x_{t} \in \mathcal{X}$. This translates into an affine constraint on $\bu_t$, since the control action mush satisfy 
\begin{equation*} 
H(\x_t) \bu_t \in \mathcal{X}-f(\x_t). 
\end{equation*}
Since $x_t$ is known, $H(x_t)$ is a constant in the above equation, and the constraint on $\bu_t$ is again polytopic. We denote this polytope as $\mathcal{S}_t$.  The set to which $\bu_t$ must belong is the intersection of $\mathcal{S}_t$ and the actuator constraints:
\begin{equation} \label{eqn:Ut} 
\mathcal{U}_t = \mathcal{S}_t \cap \mathcal{U}.
\end{equation} 
After identifying the vertices of $\mathcal{U}_t$, the algorithm in Fig.~\ref{fig:flowchart} can be used to find the optimal feedback policies. 

In general, it is fairly straightforward to find either the convex hull or the half-space representations of $\mathcal{S}_t$, since it just requires a linear transformation of $\mathcal{X}$. However, the intersection step in \eqref{eqn:Ut} and the process of finding the representation of its convex hull are non-trivial~\cite{blanchini2008set}. Below, we work through a simple example to illustrate the steps and then discuss how to overcome the computational challenges. 

\begin{figure}[!t]
	\begin{center}
		\includegraphics[width=0.95\columnwidth]{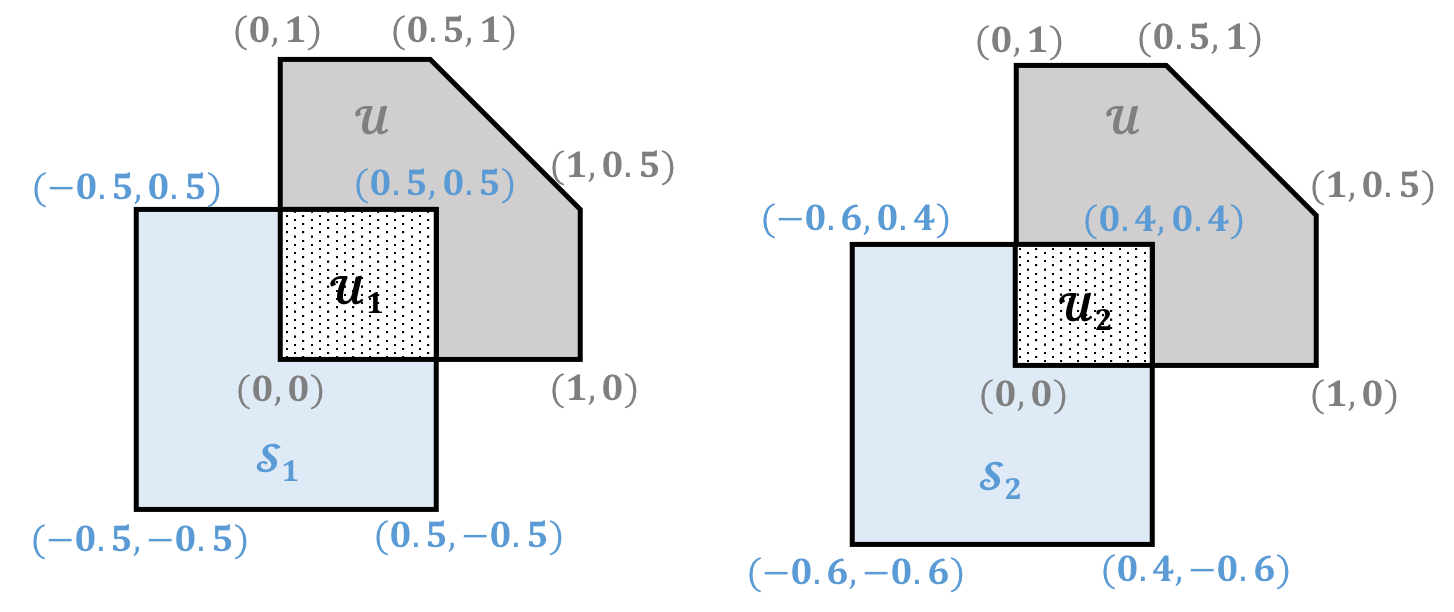}
	\end{center}
	\caption{Evolution of action safety set for a two-dimensional linear system toy example. The left plot visualizes the safety set at time $t=1$, and the right plot shows the safety set at time $t=2$.} 
	\label{fig:safety_region}
\end{figure}

\begin{example}[Intersection Step]
	Consider the following two--dimensional linear system:  \[\x_{t+1} = \begin{bmatrix} 1 & 0 \\ 0 & 1\end{bmatrix}\x_t + \begin{bmatrix} 1 & 0 \\ 0 & 1\end{bmatrix}\bu_t.\] Suppose the action safety set $\mathcal{U}$ is a convex polytope defined by: $0 \leq u_1 \leq 1$, $0 \leq u_2 \leq 1$ and $u_1+u_2 \leq 1.5$. The state safety set $\mathcal{X}$ is a square defined by $0 \leq x_1 \leq 1, 0 \leq x_2 \leq 1$ and the initial state is $x_0 = \begin{bmatrix} 0.5 & 0.5\end{bmatrix}^T$. By simple calculation, $\mathcal{S}_1 = \{-0.5 \leq u_1 \leq 0.5, -0.5 \leq u_2 \leq 0.5\}$ and $\mathcal{U}_1$ is the box bounded by $\{(0,0), (0, 0.5), (0.5, 0), (0.5, 0.5)\}$. 
	Fig.~\ref{fig:safety_region} (left) visualizes the intersection operations. 
	
	Now suppose a feasible action $u = [0.1\ 0.1]^\top$ is chosen and the system evolves. Then, $\mathcal{S}_2 = \{-0.6 \leq u_1 \leq 0.4, -0.6 \leq u_2 \leq 0.4\}$. Performing the intersection of $\mathcal{S}_2$ and $\mathcal{U}$, we get that $\mathcal{U}_2$ is a rectangle defined by the vertices $\{(0,0), (0, 0.4), (0.4, 0), (0.4, 0.4)\}$ as depicted in Fig.~\ref{fig:safety_region} (right).
\end{example}  

\subsection{Intersection of Polytopes}
It should be noted that finding the vertices of an intersection of polytopes is not easy~\cite{tiwary2008hardness}. If the polytopes are in half-space representation, then their intersection can be found by simply stacking the inequalities. However, finding the vertices of the resulting polytope can be computationally expensive. Similarly, directly intersecting two polytopes based on their convex hull representation is also intractable in general. 

Luckily, in many applications, we are not intersecting two generic polytopes at each step. Rather, there are only two "basic" sets, $\mathcal{X}$ and $\mathcal{U}$, and we are intersecting a linear transformation of these. It turns out that for many systems (see Section~\ref{sec:simulation}), we can find the resulting vertices by hand-designed rules. In addition, there are heuristics that work well for low-dimensional systems~\cite{broman1990compact}. Applying the proposed VN  technique to high-dimensional systems is the main future direction for this work.  

In the case that $\mathcal{S}_t$ and $\mathcal{U}$ do not overlap, one can choose to stop the training process. However, in our rules of finding vertices, we pick the point in $\mathcal{U}$ that closest to $\mathcal{S}_t$ to be the vertex. By design, the output of the VN is the action within set $\mathcal{U}$, meanwhile transiting to the state closest to the safe state set $\mathcal{X}$.

\subsection{Safe layer}
\label{sec:safe_layer}
Once we obtain $\mathcal{U}_t$, the next step is to encode the geometry information into the policy network such that the generated action stays in $\mathcal{U}_t$. According to Proposition~\ref{lem:vertices}, it suffices for the policy network to generate the weights (or coefficients) of that convex combination. 

\begin{figure}[h]
	\centering   
	\includegraphics[width=0.95 \columnwidth]{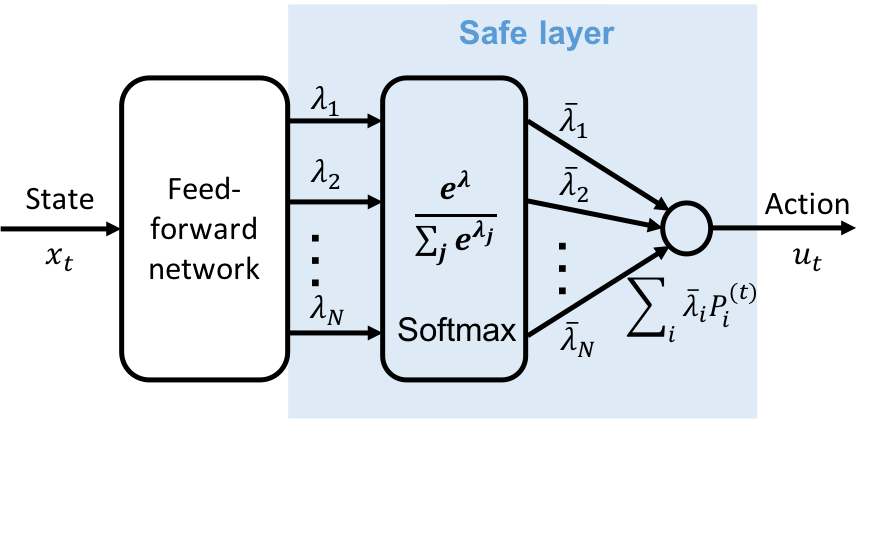}
	\caption{Illustration of the proposed safe layer architecture. The output of the policy network is modified to predict the weights $\lambda_i, i \in [1, N]$. These weights are normalized to $\bar{\lambda}_i, i \in [1, N]$ that satisfies $\bar{\lambda}_i \geq 0$ and $\sum_{i=1}^{N} \bar{\lambda}_i = 1$, via the softmax activation function. The action output is calculated as $\sum_{i=1}^{N} \bar{\lambda}_i P_i^{(t)}$, where $P_i^{(t)}$ are the safety polytope vertices. }
	\label{fig:safe_layer}
\end{figure}

Suppose that $\mathcal{U}_t$ can have at most $N$ vertices, labeled $P_1^{(t)},\ldots,P_N^{(t)}$. In the policy network architecture design, we add an intermediate safe layer that first generates $N$ nodes $\lambda_1, \ldots, \lambda_N$. The value of these nodes, however, are not positive nor do they sum to $1$. Therefore, a softmax unit is included as the activation function in order to guarantee the non-negativity and the summation constraints. In particular, we define $\bar{\lambda}_i=e^{\lambda_i}/(\sum_{j=1}^N e^{\lambda_j})$, the weights of a convex combination. The final output layer (action $u_t$) is defined as the multiplication of these normalized weights $\bar{\lambda}_i$ and the corresponding vertex values,
\begin{equation} 
\bu_t = \sum_{i=1}^N \bar{\lambda}_i P_i^{(t)}.
\end{equation} 
An illustration diagram is provided in Fig.~\ref{fig:safe_layer}.

\section{Simulation}
\label{sec:simulation}
In this section, we present and analyze the performance of the proposed VN. We  first describe the baseline algorithms and then demonstrate the performance comparisons in three benchmark control tasks: (i) inverted pendulum, (ii) mass-spring and (iii) hovercraft tracking.

\subsection{Baseline Algorithm and Architecture Design}

As mentioned earlier, the baseline algorithm for the policy update is DDPG~\cite{lillicrap2015continuous}, which is a state-of-the-art continuous control RL algorithm. To add safety constraints to the vanilla DDPG algorithm, a natural approach is to artificially shape the reward such that the agent will learn to avoid undesired areas. This can be done by setting a negative reward to the unsafe states and actions. In standard policy network (PN) baselines to which we compare, we include such a soft penalty in the rewards. We train such PN along with VN for comparison. The main difference between PN and VN is that PN only has the feed-forward network (white block in Fig.~\ref{fig:safe_layer}) and does not contain the final safe layer (blue block). The output of PN is truncated to ensure the actuator safety constraints. 

We use the following hyperparameters for all experiments. For PN, we use a three-layer feed-forward neural network, with $256$ nodes in each hidden layer. For VN, it has two feed-forward layers (with $256$ nodes in each hidden layer) and a final safe layer as described in Section~\ref{sec:safe_layer}.

\subsection{Pendulum}

For the inverted pendulum simulation, we use the OpenAI gym environment ({\tt pendulum-v0}), with the following pendulum specifications: mass $m=1$, length $l=1$. The system state is two-dimension that include angle $\theta$ and angular velocity $\omega$ of the pendulum, and the control variable is the applied torque $u$. We set the safe region to be $\theta \in [-1, 1]$ (radius) and torque limits $u \in \mathcal{U} = [-15, 15]$. The reward function is defined as $r = -(\theta^2 + 0.1 \omega^2 + 0.001 u^2)$, with the goal of learning an optimal feedback controller. 

With a discretization step size of  $\Delta = 0.05$, the following are the discretized system dynamics: 
\begin{align}
\theta_{t+1} & = \theta_t + \omega_t \Delta + \frac{3g}{2l} \sin(\theta_t) \Delta^2 + \frac{3}{ml^2} u \Delta^2 \label{eq:pendulum_dynamics} \\
\omega_{t+1} & = \omega_t + \frac{3g}{2l} \sin(\theta_t) \Delta + \frac{3}{ml^2} u \Delta \notag
\end{align}

To keep the next state in the safe region $\theta_{t+1} \in [-1, 1]$, we can compute the corresponding upper and lower bound of $u$ to represent set $\mathcal{S}_t$ by \eqref{eq:pendulum_dynamics}. Therefore, the vertices of VN can be found by intersecting $\mathcal{S}_t$ and $\mathcal{U}$. Under the case where $\mathcal{S}_t$ and $\mathcal{U}$ have no overlap, we pick $-15$ as the vertices if the upper bound of $\mathcal{S}_t$ is less than $-15$. Otherwise, we pick $15$ as the vertices.
\begin{figure}[h]
	\begin{center}
		\includegraphics[width=\columnwidth]{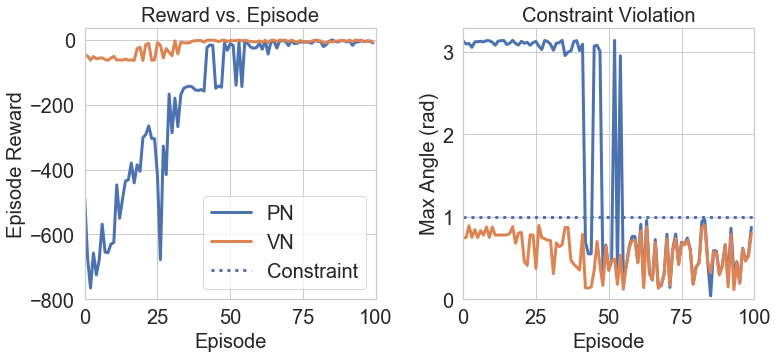}
	\end{center}
	\caption{Comparison of accumulated reward and constraint violation (max angle) for the pendulum problem using PN and VN.}
	\label{fig:pendulum_reward_constraint}
\end{figure}

For comparison, the output of PN is constrained in $[-15, 15]$ using ${\tt tanh}$ activation function in the final layer. The initial state of each episode is randomly sampled in the safe state region $[-1, 1]$. In Fig.~\ref{fig:pendulum_reward_constraint}, we show a comparison of the accumulated reward and the max angle of each episode in training of PN and VN. We observe that VN maintains safety throughout the training process, and as a result, it achieves higher reward in the early stage. It is also interesting to observe that the PN also becomes ``safe'' after training, since the reward function itself drives $\theta$ to be small. This suggests if we can train the PN offline, it might be able to obey the safety constraint for some control systems. However, the next example shows that even a well-trained policy may violate constraints if these hard constraints are not explicitly taken into account. 


\subsection{Mass-Spring}

Now we consider the task of damping an oscillating system Mass-Spring to the equilibrium point \cite{blanchini2008set}. The system includes a mass $m=1$ and a spring $k=1$ and the state is two-dimensional with position $x$ and speed $v$ of the mass. The control variable is the force exerted on the mass $u$. We set the safe region to be $v \in [-1, 1], u \in [-1, 1]$. We define the reward function to be $r = -(x^2 + v^2) $. The initial state are randomly sampled from $x\in [-2, 2], v \in [-1, 1]$.

The system dynamics are defined as follows,
\begin{equation*}
\begin{bmatrix}x_{t+1}\\ v_{t+1}\end{bmatrix}  =\begin{bmatrix}
1 & \Delta\\
-\frac{k}{m}\Delta & 1
\end{bmatrix}\begin{bmatrix}
x_t \\ v_t
\end{bmatrix} +\begin{bmatrix}
0\\
\frac{1}{m}\Delta
\end{bmatrix}u
\end{equation*}


\begin{figure}[h]
	\begin{center}
		\includegraphics[width=\columnwidth]{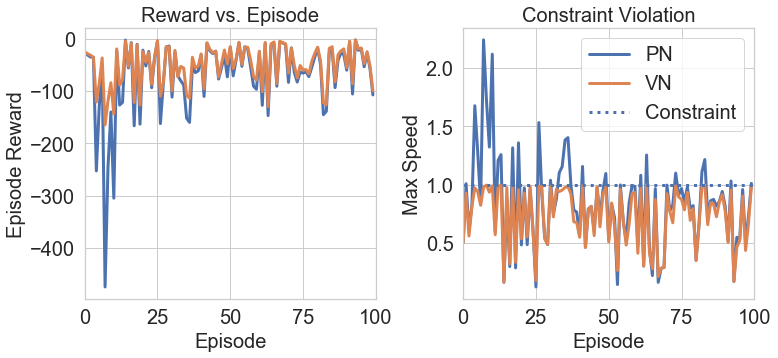}
	\end{center}
	\caption{Comparison of accumulated reward and constraint violation (max speed) for Mass-Spring problem using PN and VN.}
	\label{fig:mass_spring_reward_constraint}
\end{figure}

Fig. \ref{fig:mass_spring_reward_constraint} compares the accumulated reward and the max speed of each episode in the training of PN and VN. VN maintains safety during training and receives higher reward in the early training stage. Note that even trained PN could still violate the constraints.

\subsection{Hovercraft}
\begin{figure}[h]
	\begin{center}
		\includegraphics[width=0.4\columnwidth]{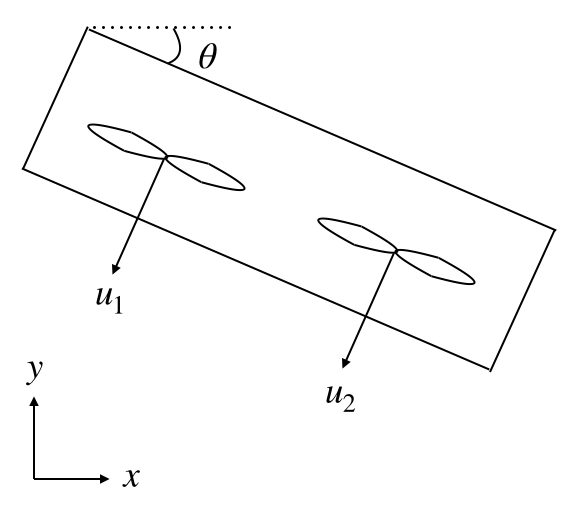}
		\includegraphics[width=0.5\columnwidth]{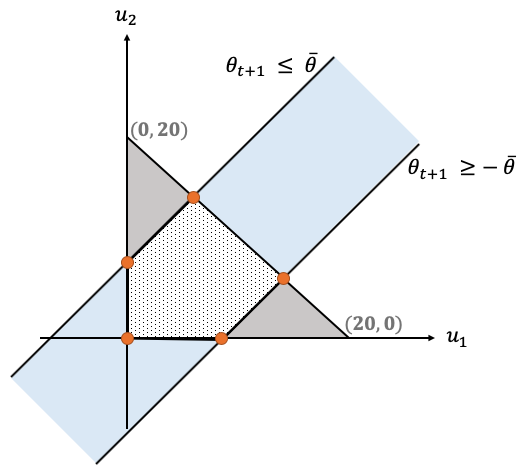}
	\end{center}
	\caption{(Left) Hovercraft example. $u_1$ and $u_2$ denote starboard and port fan forces. $\theta, x, y$ are the tilt angle and the coordinate position. (Right) Illustration of computing vertices of intersection of polytopes $\mathcal{S}_t$ and $\mathcal{U}$.}
	\label{fig:hovercraft}
\end{figure}

Consider the task of controlling a hovercraft illustrated  in Fig.~\ref{fig:hovercraft} (left), which has system dynamics defined as follows:

\begin{align*}
x_{t+1} & = x_{t} + v_{x, t}\Delta  + \frac{1}{2m} \sin \theta_t (u_1 + u_2) \Delta^2\\
v_{x, t+1} & = v_{x, t} + \frac{1}{m} \sin \theta_t (u_1 + u_2) \Delta \\
y_{t+1} & = y_{t} + v_{y, t}\Delta  + \frac{1}{2m} (\cos \theta_t (u_1 + u_2) - g) \Delta^2\\
v_{y, t+1} & = v_{y, t} + \frac{1}{m} (\cos \theta_t (u_1 + u_2) - g) \Delta \\
\theta_{t+1} & = \theta_t + v_{\theta, t} \Delta  + \frac{1}{2l} (u_1 - u_2) \Delta^2 \\
v_{\theta, t+1} & = v_{\theta, t} + \frac{1}{l} (u_1 - u_2) \Delta
\end{align*}

Let $m = l = 1, g =10$. Considering the force exerted on two fans are coupled, the actuator constraint set is defined as $\{u_1, u_2 | u_1 \ge 0, u_2 \ge 0, u_1 + u_2 \le 20 \}$. 
To keep the tile angle of the hovercraft in a safety region, we set the safe state region to be $\theta \in [-\overline{\theta}, \overline{\theta}]$. Define the reward function $r = -(x - x_0)^2 - (y - y_0)^2 - \theta^2 - 0.1 (v_x^2 + v_y^2 + v_\theta^2) - 0.001 (u_1^2 + u_2^2)$ to learn a controller that tracks the target position $(x_0, y_0)$. Fig.~\ref{fig:hovercraft} (right) shows how to use at most five vertices to represent the intersection of safe state region and the actuator constraint region.

\begin{figure}[h]
	\begin{center}
		\includegraphics[width=\columnwidth]{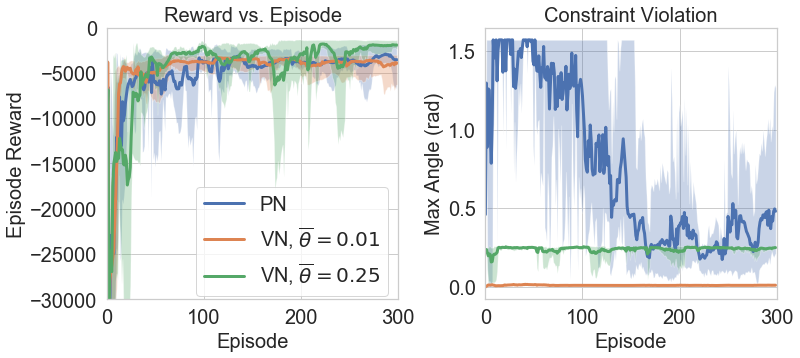}
	\end{center}
	\caption{Comparison of accumulated reward and constraint violation (max tile angle) from Hovercraft control using PN and VN with different constraint upper bound.}
	\label{fig:hovercraft_reward_constraint}
\end{figure}

\begin{figure}[h]
	\begin{center}
		\includegraphics[width=\columnwidth]{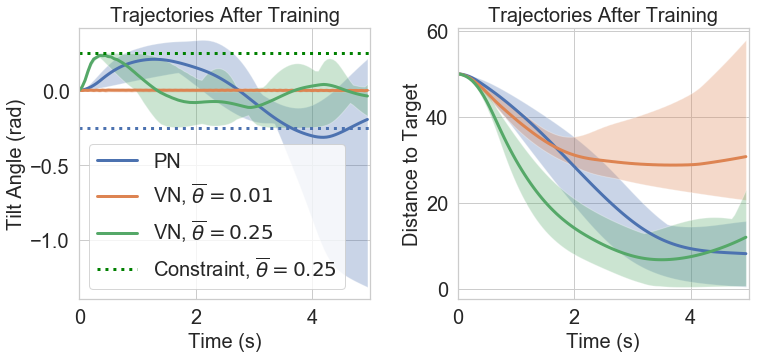}
	\end{center}
	\caption{Trajectories generated by trained PN and VN (with different tilt angle upper limit) policies. (Left) tilt angle and (Right) square of distance to the target position.}
	\label{fig:hovercraft_policy}
\end{figure}
In our experiment, the initial state is set at position $(0, 0)$ and the target position is $(5, 5)$. 
To better investigate the effect of the constraint, we train VNs for tilt angle upper bounds of $\overline{\theta} = 0.01$ and $\overline{\theta} = 0.25$ radians. 
Fig. \ref{fig:hovercraft_reward_constraint} compares the accumulated reward and max tilt angle of each episode in the training of PN and VN. Fig. \ref{fig:hovercraft_policy} visualizes the trajectories of trained PN and VN policies. In the trajectories, we observe that the angle of hovercraft first turn positive to have some momentum pointed to the right, then turn to slow down the speed. When $\overline{\theta} = 0.01$, the hovercraft has a strict constraint on its tilt angle and is unable to reach the target position. In both choices of the tilt angle upper limit $\overline{\theta}$, the constraint is never violated in the whole trajectory executing learned VN. However, running learned PN will still reach large tilt angle, even if the soft penalty is added in the reward.


\section{Conclusions}
Motivated by the problem of training an RL algorithm with hard state and action constraints, leveraging
the geometric property that a convex polytope can be equivalently represented as the convex hull of a finite set of vertices, we design a novel policy network architecture called VN that guarantees the output satisfies the safety constraints by design. Empirically, we show that VN yields significantly better safety performance than a vanilla policy network architecture with a constraint violation penalty in several benchmark control systems. An important future direction is to extend the proposed method to high-dimensional control systems.





\bibliographystyle{apalike}
\bibliography{refs}

\end{document}